\documentclass[journal]{IEEEtran}
\usepackage[usenames]{color}
\usepackage{epsfig}
\usepackage{graphics}
\usepackage{cite}
\usepackage{array}
\usepackage{pslatex} 
\usepackage{url}
\usepackage{caption}
\usepackage{lineno}
\usepackage{amsmath,amssymb,amsfonts}
\usepackage{algorithmic}
\usepackage{graphicx}
\usepackage{subfig}
\usepackage{textcomp}
\usepackage{xcolor}
\usepackage{colortbl}%
\usepackage{multirow}
\usepackage{booktabs}

\usepackage{graphicx}  
\usepackage{setspace}
\usepackage{tikz}
\usepackage{hhline}
\usepackage{mathtools}
\usepackage[colorlinks=true, linkcolor=blue, citecolor=blue, urlcolor=blue, pdfborder={0 0 0}]{hyperref}

\usepackage{bbm}
  
\newcolumntype{C}[1]{>{\centering\arraybackslash}m{#1}}

\def\BibTeX{{\rm B\kern-.05em{\sc i\kern-.025em b}\kern-.08em
    T\kern-.1667em\lower.7ex\hbox{E}\kern-.125emX}}
    
\usepackage[letterpaper]{geometry}
 \newcommand{\flogo}{\includegraphics[height=18pt]{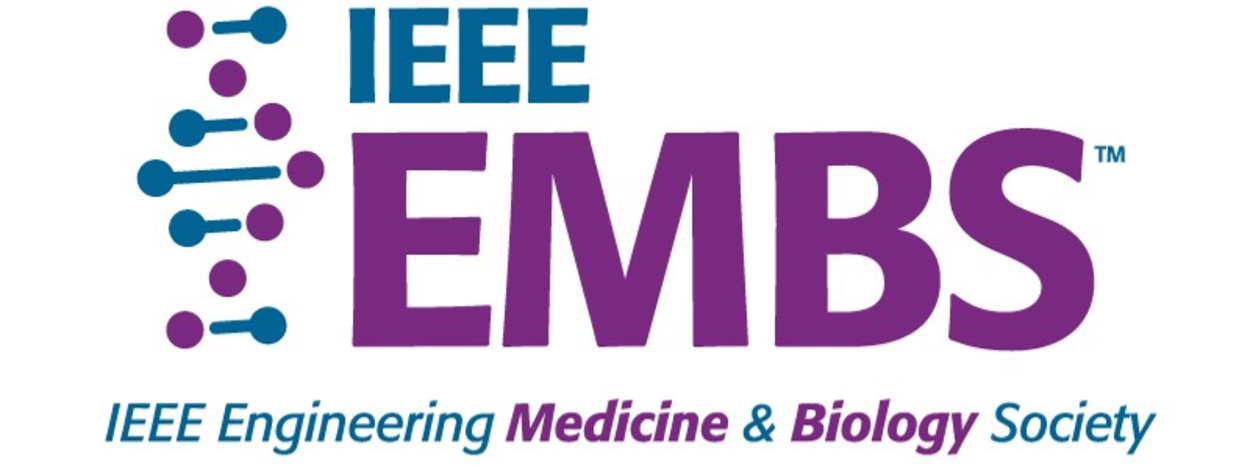}
 }
\newcolumntype{C}[1]{>{\centering\arraybackslash}p{#1}}

\ifCLASSINFOpdf
\else
\fi
\usepackage[english]{babel}
\usepackage[utf8]{inputenc}
\usepackage{fancyhdr}
\newcommand{\figref}[1]{Fig.~\ref{fig:#1}}
\newcommand{\secref}[1]{Section~\ref{sec:#1}}
\newcommand{\tblref}[1]{Table~\ref{tab:#1}}

\begin{document}

\title{\textcolor{violet}{LLM-Powered Counterfactuals for Health Intervention Design and Sensor Data Augmentation
\vspace{0.25cm}}
}
\title{\textcolor{violet}{Counterfactual Modeling with Fine-Tuned LLMs for Health Intervention Design and Sensor Data Augmentation
\vspace{0.25cm}}
}

\author{
Shovito Barua Soumma$^{1,2}$,~\IEEEmembership{Student Member,~IEEE}, 
Asiful Arefeen$^{1,2}$, 
Stephanie M. Carpenter$^{1}$, 
Melanie Hingle$^{3}$, 
and Hassan Ghasemzadeh$^{1}$,~\IEEEmembership{Senior Member,~IEEE}
\thanks{
$^{1}$College of Health Solutions, Arizona State University, Phoenix, AZ 85004, USA. 
Emails: \{shovito, aarefeen, stephanie.m.carpenter, hghasemz\}@asu.edu.
}
\thanks{
$^{2}$School of Computing and Augmented Intelligence, Arizona State University, Tempe, AZ 85281, USA.
}
\thanks{
$^{3}$School of Nutritional Sciences and Wellness, University of Arizona, Tucson, AZ 85721, USA. 
Email: hinglem@arizona.edu.
}
\thanks{
This work was supported in part by the National Science Foundation (NSF) under Grant IIS-2402650. 
A. Arefeen was supported in part by the National Institute of Diabetes and Digestive and Kidney Diseases (NIDDK) of the National Institutes of Health (NIH) under Award T32DK137525. 
The content is solely the responsibility of the authors and does not necessarily represent the official views of the NSF or NIH.
}
%
}

%



\maketitle\thispagestyle{fancy}

\begin{abstract}
Counterfactual explanations (CFEs) provide human-centric interpretability by identifying the minimal, actionable changes required to alter a machine learning model’s prediction. Therefore, CFs can be used as (i) interventions for abnormality prevention and (ii) augmented data for training robust models. we conduct a comprehensive evaluation of CF generation using large language models (LLMs), including GPT-4 (zero-shot and few-shot) and two open-source models—BioMistral-7B and LLaMA-3.1-8B—in both pretrained and fine-tuned configurations. Using the multimodal AI-READI clinical dataset, we assess CFs across three dimensions: intervention quality, feature diversity, and augmentation effectiveness. Fine-tuned LLMs, particularly LLaMA-3.1-8B, produce CFs with high plausibility (up to 99\%), strong validity (up to 0.99), and realistic, behaviorally modifiable feature adjustments. When used for data augmentation under controlled label-scarcity settings, LLM-generated CFs substantially restore classifier performance, yielding an average ~20\% F1 recovery across three scarcity scenarios.
Compared with optimization-based baselines such as DiCE, CFNOW, and NICE, LLMs offer a flexible, model-agnostic approach that generates more \textcolor{black}{clinically plausible, behaviorally actionable} and semantically coherent counterfactuals. Overall, this work demonstrates the promise of LLM-driven counterfactuals for both interpretable intervention design and data-efficient model training in sensor-based digital health.
\end{abstract}

\begin{IEEEkeywords}
Counterfactual explanations, Digital health, Explainable AI, Label Scarcity, Large Language Model (LLM)
\end{IEEEkeywords}

%
\IEEEpeerreviewmaketitle

\textbf{\textit{Impact Statement-} SenseCF fine-tunes an LLM to generate valid, representative counterfactual explanations and supplement minority class in an imbalanced dataset for improving model training and boosting model robustness and predictive performance.}\\
\\
\vspace{-3mm}
\section{\textbf{INTRODUCTION}}

\IEEEPARstart{A}{ccurate} and interpretable predictions from machine learning (ML) models are increasingly vital in healthcare applications such as disease risk forecasting and sleep efficiency estimation using physiological and sensor data. While these models excel at outcome prediction, they often fall short in guiding actionable interventions to reverse adverse outcomes- specially in black-box settings~\cite{soumma2025ai}.

\textcolor{black}{Counterfactual explanations (CFEs) offer a powerful approach for model interpretability by identifying the minimal, actionable changes needed to reverse an undesirable prediction, a concept first formalized in the seminal work of Wachter et al~\cite{Wachter2017CounterfactualEW}. Subsequent surveys and methodological frameworks~\cite{Guidotti2018ASO,Karimi2020AlgorithmicRF,Karimi2019ModelAgnosticCE} have emphasized CFs’ central role in actionable recourse, feasibility constraints, and user-centered interpretability. Traditional optimization-driven CF methods such as DiCE~\cite{mothilal2020dice}, CFNOW~\cite{DEOLIVEIRA2023}, and NICE~\cite{Brughmans2021NICEAA} often require access to model gradients or internal structure, limiting their real-world applicability and frequently struggling with categorical coherence or clinically plausible modifications. In contrast, large language models (LLMs) provide a promising alternative: leveraging zero- and few-shot prompting, they can generate realistic, semantically consistent counterfactuals using only input–output context~\cite{mann2020language,soumma2025sensecf}. This paradigm not only eliminates the dependency on gradients or model access but also opens the door for scalable, human-centered explanations across diverse datasets.
}

Recent work highlights that LLMs possess strong innate counterfactual reasoning abilities even without task-specific fine-tuning~\cite{soumma2025sensecf,fizle_huan,li-etal-2024-prompting, Chen2025CounterBenchAB}, building on earlier foundations showing that counterfactual reasoning supports actionable model transparency~\cite{Wachter2017CounterfactualEW,Guidotti2018ASO,Karimi2020AlgorithmicRF,Karimi2019ModelAgnosticCE}. However, their application to structured and multimodal health data—where features arise from heterogeneous physiological, behavioral, and environmental sensors—remains largely underexplored. Beyond interpretability, counterfactual explanations also serve as label-flipping synthetic samples that generate corner cases and enhance model robustness in imbalanced medical datasets, advancing prior work demonstrating the utility of counterfactuals for recourse, robustness, and data augmentation~\cite{Russell2019EfficientSF}. When used for augmentation, CFs introduce controlled perturbations that preserve the underlying data manifold while enriching decision-boundary diversity, improving resilience in low-data clinical scenarios. Although optimization-based CF approaches have shown effectiveness in decision-support settings~\cite{mothilal2020dice,DEOLIVEIRA2023}, they often struggle with categorical constraints, nonlinear physiological dependencies, and plausible semantic adjustments—limitations that LLMs can mitigate through their contextual reasoning and stronger distributional priors~\cite{Brughmans2021NICEAA,fizle_huan}.

\begin{figure}[!htb]
\vspace{-4mm}
\centering
\includegraphics[width=0.8\linewidth, trim={0 300 700 46},clip]{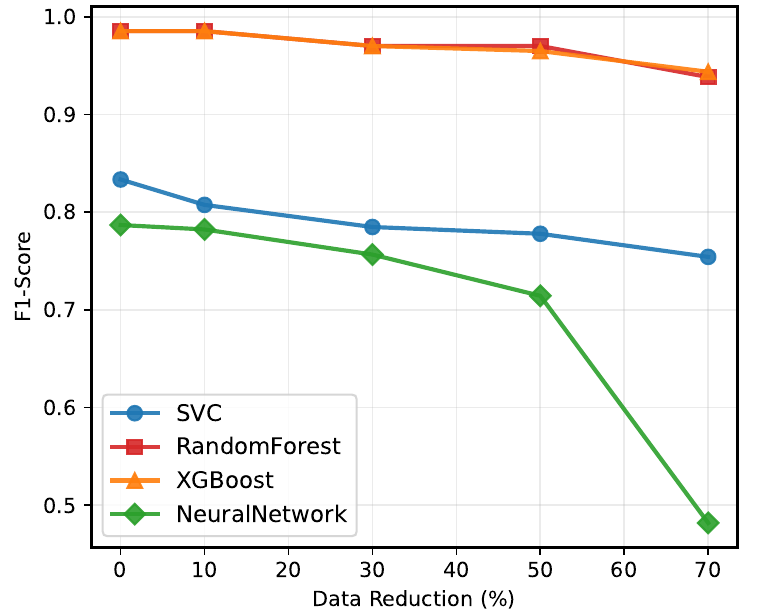} 

\caption{F1-Score Decline Across Models as Training Data is Reduced.}
\label{fig:motivation}
\end{figure}

\begin{figure*}[!htb]

\centering
\includegraphics[width=0.85\linewidth]{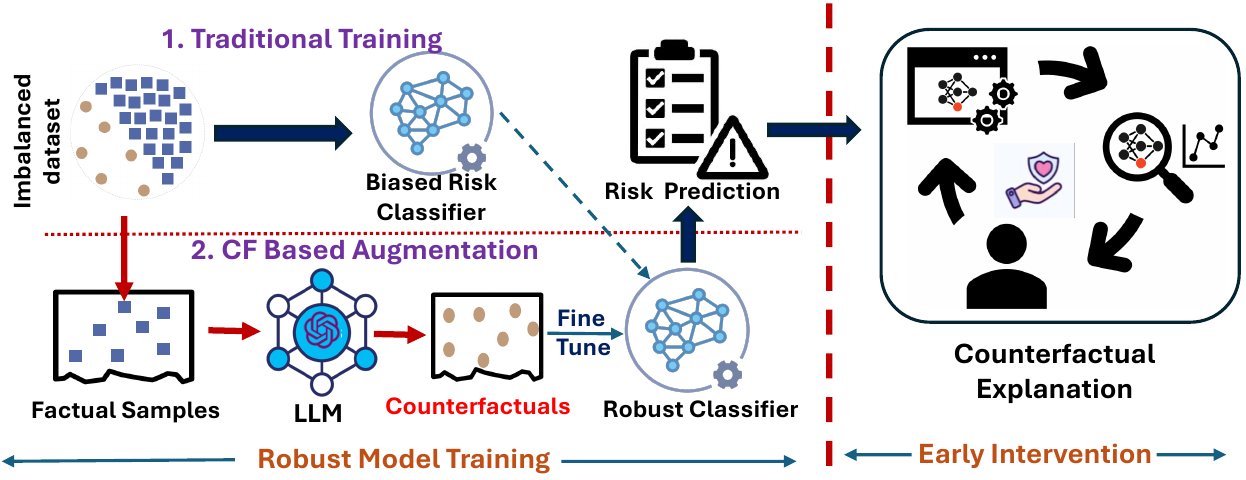} 

\caption{SenseCF pipeline: LLM-generated counterfactuals are used both for augmenting imbalanced training data (left) and for model interpretability (right).}
\vspace{-4mm}
\label{fig:method}
\end{figure*}
However, several critical gaps persist in the current literature: first, the effectiveness of LLM-based CFs have not been comprehensively evaluated on large, multimodal clinical datasets; second, standardized evaluation metrics comparing optimization-based and generative approaches remain limited; third, CFs' potential as data augmenters in healthcare scenarios remains underexplored. As illustrated in Fig.~\ref{fig:motivation}, all classifiers experience marked declines in F1-score under increasing data reduction, highlighting the vulnerability of standard models to label scarcity and motivating the need for principled synthetic augmentation via LLM-generated counterfactuals.



To address these gaps, we introduce a systematic evaluation of LLM-generated counterfactuals using zero-shot, few-shot, pretrained, and fine-tuned models across the multimodal AI-READI clinical dataset. 
This paper extends our earlier work, SenseCF~\cite{soumma2025sensecf}, which focused primarily on GPT-4o based prompting, by incorporating a broader suite of open source LLMs and performing a more comprehensive assessment of their counterfactual capabilities.
Our contributions extend beyond existing LLM-focused studies that primarily evaluate natural language processing (NLP) tasks, providing a rigorous and quantitative comparison in multimodal clinical settings~\cite{fizle_huan,li-etal-2024-prompting}. In this work, 
(i) We systematically compare GPT-4o with two open-source LLMs (BioMistral-7B and LLaMA-3.1-8B) evaluated in both pretrained and fine-tuned configurations. We assess their performance across three core dimensions: (a) actionable intervention quality, (b) feature diversity and realism, and (c) data augmentation effectiveness under controlled label-scarcity scenarios. (ii) We further benchmark their plausibility, diversity, and impact on model performance against state-of-the-art baselines. \textcolor{black}{In addition to augmentation, we treat counterfactuals as instance-level explanations, enabling interpretable and actionable insights into model predictions in clinical settings.}


\textcolor{black}{To the best of our knowledge, this is the first systematic evaluation\footnote{A preliminary version of this work has been reported~\cite{soumma2025sensecf}.}  of fine-tuned LLM-generated counterfactuals as structured augmentation samples for improving robustness under label scarcity in sensor-derived health datasets, while also assessing their utility for intervention-oriented explanation.}

\section{\textbf{MATERIALS AND METHODS}}
\begin{figure}[!h]
\vspace{-4mm}
\centering
\includegraphics[width=0.9\linewidth]{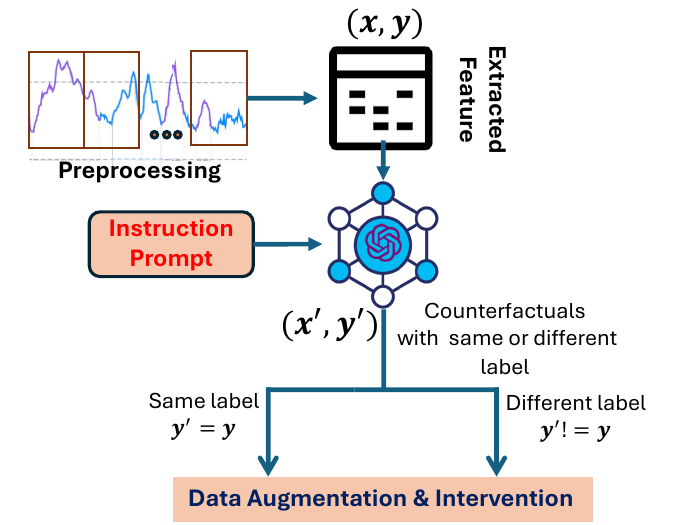}
\caption{Counterfactual generation using LLMs from sensor-derived features.}
\label{fig:llm-cf-pipeline}
\end{figure}
In this section, we detail our approach for generating and evaluating counterfactual explanations (CFs) using large language models (LLMs). As illustrated in Fig~\ref{fig:llm-cf-pipeline} and Fig~\ref{fig:method}, our methodology aims at: (1) producing actionable counterfactual (CF) interventions by reversing the predictions of trained ML models, and (2) leveraging these CFs as augmented training data to enhance model robustness, specifically under label scarcity. 

\subsection{Study Design}
We represent our input data as a set of tuples $(x_i,y_i)$, where $x_i\in X$ is a feature vector representing either clinical or physiological data and $y_i\in \{0,1\}$ denotes the ground truth label. A trained predictive model $f(\cdot)$ outputs predictions $\hat{y_i}=f(x_i)$. Our preprocessing stage (discussed in Section~\ref{sec-preprocess}) transforms raw data into structured, tabular feature vectors suitable for prompting the LLM. 

\subsubsection{Counterfactual Generation}
To generate counterfactuals, we used three LLM families under four generative modes:
\begin{enumerate}
    \item \textbf{GPT-4o: }Zero-shot (ZS) and few-shot (FS) prompting
    \item \textbf{BioMistral-7B:} Zero-shot and fine-tuned (FT) on the structured training data
    \item \textbf{LLaMA-3.1-8B:} Zero-shot and fine-tuned on the structured training data
\end{enumerate}

\begin{figure}[!htb]
\centering
\vspace{-4mm}
\includegraphics[width=0.9\linewidth]{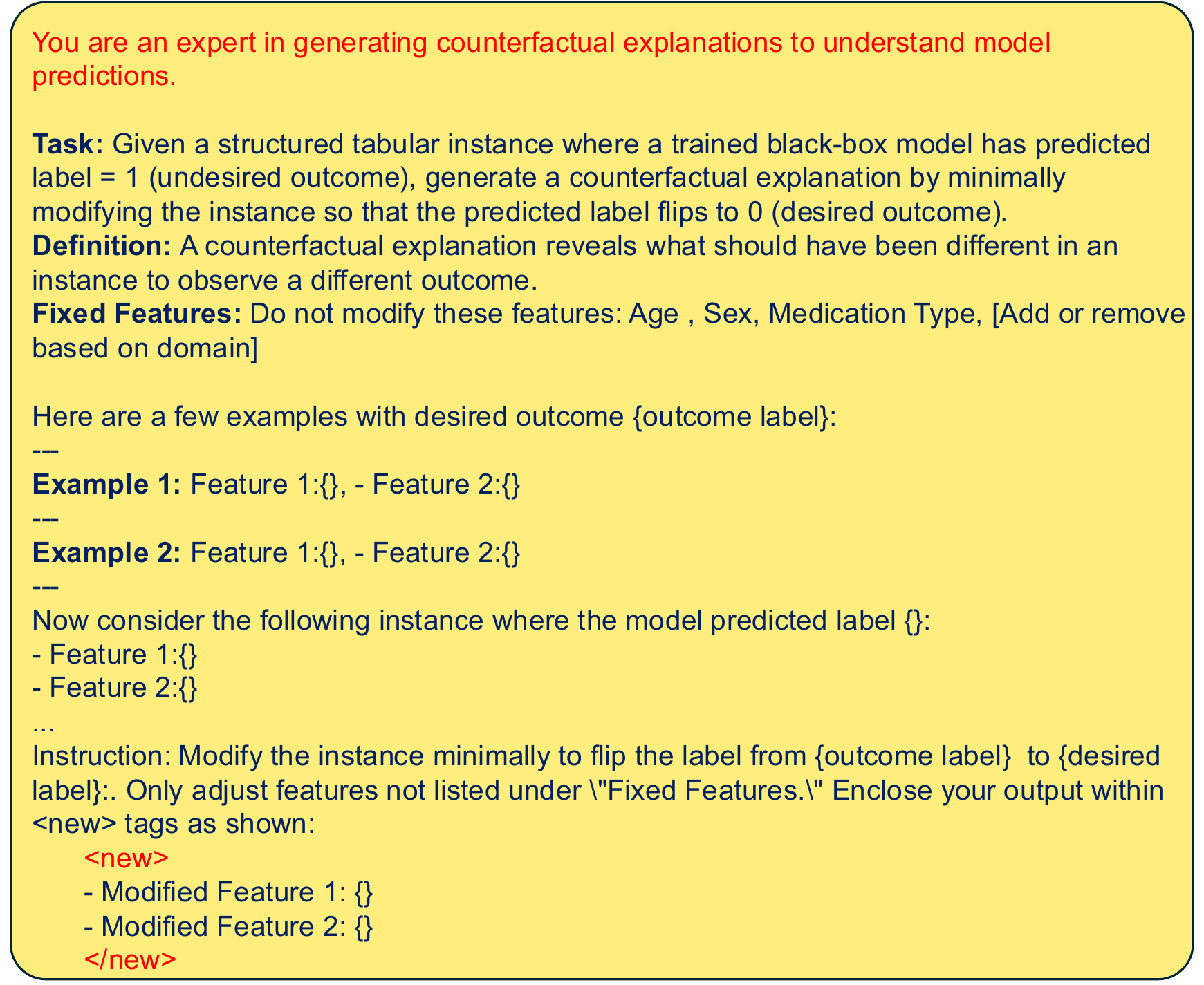}
\caption{Prompt template for counterfactual generation.}
\label{fig:llm-prompt}
\vspace{-4mm}
\end{figure}

\textbf{Problem Formulation:} Given a feature vector $x_i$ and the prediction $\hat{y_i}$, each LLM generates a modified vector $x^{'}_i$, where the model’s prediction changes from $\hat{y_i}$ to a desired opposite outcome $y_i\neq\hat{y_i}$. We also explicitly constrain the LLM from altering immutable or clinically fixed features (e.g., age, sex, or medication type), ensuring that generated counterfactuals remain actionable and plausible within domain constraints. The generation of CFs can be described as:
\[
x'_i = \text{LLM}(x_i, \text{prompt}), \quad \text{s.t. } f(x'_i) \neq f(x_i)
\]
The instructional prompt explicitly constrains LLM to minimally alter feature values to achieve a realistic and actionable counterfactual, ensuring the plausibility and feasibility of generated CFs. This multi-model setup enables us to compare the semantic precision, boundary awareness, and distributional alignment of counterfactuals produced by proprietary vs. open-source models and by zero-shot prompting vs. supervised fine-tuning.
\textcolor{black}{
To convert LLM-generated counterfactuals into structured feature vectors, we enforce a predefined output format using \texttt{<new>} tags (\figref{llm-prompt}). A rule-based parser extracts modified feature--value pairs and maps them onto the original feature vector $x_i \in \mathbb{R}^d$. The counterfactual $x_i'$ is constructed as:
}
\begin{equation}
\color{black}
x_{i,j}' =
\begin{cases}
\tilde{x}_{i,j}, & \text{if feature } j \text{ is modified by the LLM} \\
x_{i,j}, & \text{otherwise}
\end{cases}
\end{equation}
\textcolor{black}{
where $\tilde{x}_{i,j}$ denotes the parsed value from the LLM output. Immutable features $j \in F_{\text{fixed}}$ are enforced as:}
\begin{equation}
\color{black}
x_{i,j}' = x_{i,j}, \quad \forall j \in F_{\text{fixed}}.
\end{equation}
\textcolor{black}{
Additionally, all values are validated to lie within the empirical data distribution, $x_{i,j}' \in \mathrm{range}(X_j)$. This process yields a complete structured counterfactual $x_i'$, enabling consistent computation of sparsity, distance, and plausibility metrics.
}
\subsubsection{Intervention, Diversity and Data Augmentation}
Generated CFs serve three purposes in our analysis:

\textbf{(i) Intervention:} We assess whether CFs provide actionable and clinically meaningful adjustments to move a sample from an undesired label (e.g., stressed, disease-positive) to a desirable outcome. Validity, sparsity, and plausibility metrics quantify intervention realism.

\textbf{(ii) Diversity:} We also analyze how different LLMs modify feature distributions, highlighting whether fine-tuned models produce semantically grounded and structurally consistent CFs.

\textbf{(iii) Data Augmentation:}
CFs that successfully alter the model prediction are added to the training dataset to form an augmented set:
\[
X_{\text{aug}} = X \cup \{(x'_i, y'_i) \mid f(x'_i) \neq f(x_i)\}
\]

We evaluate the impact of CF augmentation under three different label-scarcity settings (Positive-Class Scarcity, Negative-Class Scarcity, Dual-Class Scarcity) and across multiple augmentation ratios. This allows us to quantify how CF quantity and LLM fine-tuning quality improve downstream classifier performance.

\subsection{Dataset}

\textit{\textbf{AIREADI data: }}
SenseCF utilizes the publicly available AIREADI Flagship Dataset~\cite{ai2024ai,Baxter2024AIREADIRA}, a resource developed to support artificial intelligence and machine-learning research in Type 2 Diabetes Mellitus (T2DM). The dataset contains information from 1,067 participants recruited across three U.S. locations: the University of Alabama at Birmingham (UAB), the University of California San Diego (UCSD), and the University of Washington (UW). AIREADI includes individuals both with and without T2DM, with careful balancing across demographic groups and diabetes status. Participants fall into four categories: healthy controls, individuals with prediabetes, individuals with T2DM treated with oral agents, and individuals with T2DM using insulin.

A major strength of the dataset is its rich multimodal composition. Each participant was monitored over a ten-day period using multiple wearable sensors: a Dexcom G6 continuous glucose monitor for frequent glucose measurements, a Garmin Vivosmart 5 for physical activity and heart-rate-variability–derived stress indices, and an Anura environmental sensor capturing factors such as air quality and temperature. Additional components include self-reported surveys, clinical examinations, and retinal images. Daily step counts were obtained via an accelerometer, with occasional missing values due to device charging.

\subsection{Preprocessing and Feature Extraction}
\label{sec-preprocess}
\begin{figure}[!h]
\vspace{-4mm}
\centering
\includegraphics[width=1\linewidth, trim={250 440 177 160},clip]{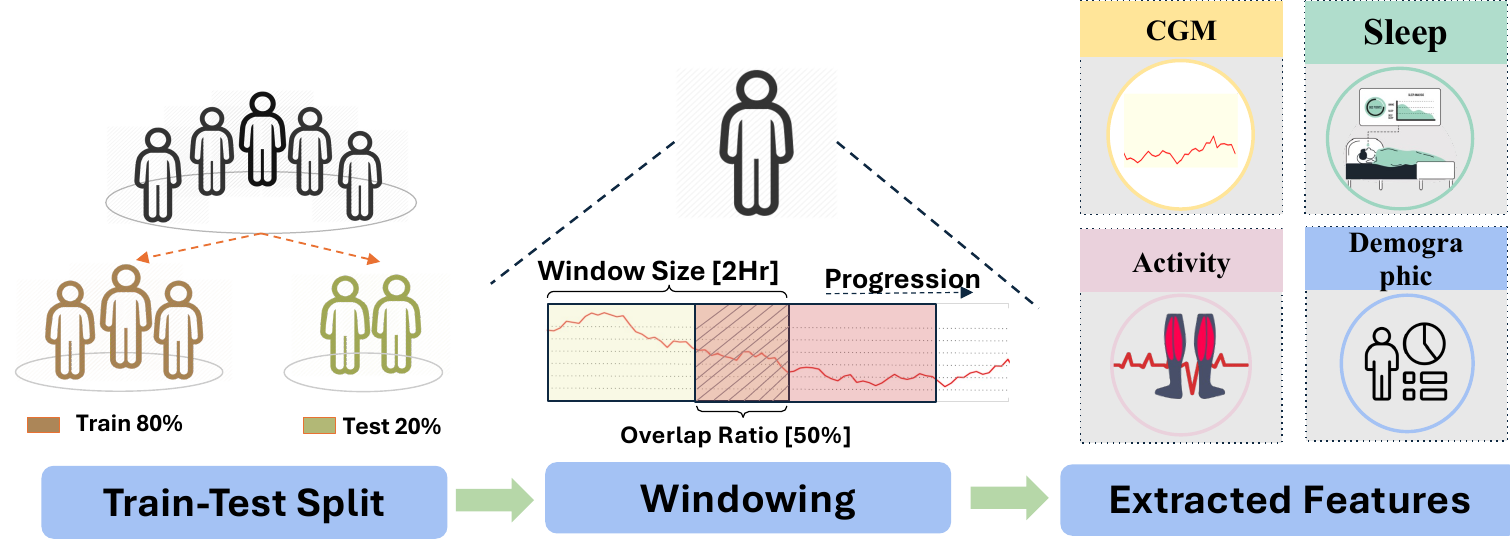}
\caption{Overview of the feature extraction pipeline. Patients are first split into train and test sets to avoid data leakage. 
    A two-hour sliding window with 50\% overlap is applied to continuous signals, followed by extraction of sleep, CGM, activity, and demographic features.}
\label{fig:preprocess_pipeline}
\vspace{-2mm}
\end{figure}
Figure \ref{fig:preprocess_pipeline} illustrates the overall preprocessing pipeline used in this study. To avoid data leakage across samples from the same individual, we perform the train–test split at the patient level, assigning 80\% of patients to the training set and the remaining 20\% to the test set. This ensures that no temporal windows, physiological patterns, or behavioral characteristics from a given patient appear in both sets, thereby preserving the integrity of model evaluation.

From each night, the \textit{Awake}, \textit{Light Sleep}, \textit{Deep Sleep} and \textit{REM Sleep} percentages are computed. A two-hour long moving window is used to extract \textit{mean step counts}, \textit{mean glucose level}, \textit{number of hyperglycemic events}, percentage of time-in-range for glucose levels \textit{(\%TIR)}. Based on high (mean daily stress $>75$) and moderate stress (mean daily stress $75$) levels ~\cite{GarminStressFAQ}, the target label is set. Therefore, a total of 12 features, including 4 immutable features: \textit{Age}, \textit{Gender}, \textit{Medication} and \textit{Patient Subgroup}, are used to classify a sample for high/moderate stress and generate the CFEs for supplementing the minority class. 

\subsection{Experimental Setup}
All experiments were conducted using a single NVIDIA A100 GPU (40 GB), which provides sufficient memory bandwidth and tensor-core acceleration for finetuning 7–8B parameter large language models. We fine-tuned two models: Llama-3.1-8B-Instruct and BioMistral-7B, using parameter-efficient LoRA adaptation rather than full-model training to reduce computational cost and prevent catastrophic forgetting of medical knowledge. Both models were trained for 2 epochs, as we found this to provide the best trade-off between learning stability and overfitting; additional epochs tended to yield diminishing returns on validation loss.
\begin{table}[h]
\centering
\small
\caption{Training Configuration for LLM Finetuning}
\begin{tabular}{l|l}
\toprule
\textbf{Component} & \textbf{Setting} \\
\midrule
GPU & NVIDIA A100\\
Models & Llama-3.1-8B, BioMistral-7B \\
Finetuning Strategy & LoRA (PEFT) \\
LoRA Rank ($r$) & 16 \\
LoRA Alpha & 32 \\
LoRA Dropout & 0.03 \\
Quantization & 4-bit NF4 (BitsAndBytes) \\
Epochs & 2 \\
Batch Size & 4 (effective 16 with GA=4) \\
Learning Rate & 2e-4 \\
Max Sequence Length & 512 tokens \\
Precision & FP16 \\
Eval Frequency & Every 50 steps \\
Checkpoint Frequency & Every 100 steps \\
\bottomrule
\end{tabular}
\label{tab:training_setup}
\end{table}
We applied 4-bit NF4 quantization (BitsAndBytes) only to the Llama-3.1-8B model in order to substantially reduce its memory footprint on the A100 GPU. LLaMA models have larger attention projections and higher activation memory compared to BioMistral-7B, making them more memory-intensive during finetuning. Quantizing LLaMA to NF4 allows the model to fit comfortably in GPU memory, enables larger effective batch sizes, and avoids out-of-memory (OOM) errors during training—while preserving model quality and maintaining stable optimization under LoRA.
\subsection{Baselines}

We have identified the following techniques to compare against SenseCF.

 \textit{\textbf{(i) DiCE}} \cite{mothilal2020dice} generates a diverse set of CFs to maximize variability across solutions while also optimizing for proximity, and feasibility across local regions of the decision boundary.

    \textit{\textbf{(ii) CFNOW}} \cite{DEOLIVEIRA2023} is a model-agnostic method that employs a two-step search algorithm to explore the search space and generate valid and minimal CFs.
    
    \textit{\textbf{(iii) NICE}}~\cite{Brughmans2021NICEAA} iteratively constructs CFs by replacing feature values with those from the nearest instance having a different prediction

\subsection{Evaluation Metrics}
We assess the CFs using some standard metrics found in the literature:

\textbf{\textit{Validity}} assesses whether the produced CFs genuinely belong to the desired class \cite{Hamman2023RobustCE}. High validity indicates the technique’s effectiveness in generating valid CF examples.
\begin{equation*}
    \textit{validity} = \frac{\#|f(X_T^*) \neq f(X_T)|}{\|CF\|}
\end{equation*}

\textbf{\textit{Distance}} between the CF and the factual sample is calculated from the $L_2$ normalized distance of the continuous features and the hamming distance of the categorical features \cite{Karimi2020AlgorithmicRF}.

\textbf{\textit{Sparsity}} is the average number of feature changes per CF \cite{Guo2021CounterNetET}. A low sparsity ensures better user understanding of the CFs.
\begin{equation}
\textit{sparsity} = \frac{\sum_{X_T^*\in CF}^{}\sum_{i=1}^{d} \mathbbm{1}(x_T^{*i} \neq x_T^i)}{\|CF\|} 
\end{equation}

\textbf{\textit{Plausibility}} quantifies the fraction of explanations that fall within the feature ranges derived from the data \cite{Guidotti2022CounterfactualEA}-
\begin{equation*}
    \textit{plausibility}=\frac{\sum_{X_T^*\in CF}^{}\mathbbm{1}(\text{dist}(X_T^*)\subseteq\text{dist}(X))}{\|CF\|}
\end{equation*}
where, dist($X_T^*$) and dist($X$) represent the distribution of feature values in the CF instances $X_T^*$ and in the training data, respectively. $\|CF\|$ is the total number of CF instances.

\section{\textbf{RESULTS}}
Our evaluation highlights the dual role of LLM-generated CFs as highly plausible interventions and as impactful data augmenters for robust model training in digital health contexts.
\subsection{Intervention \textcolor{black}{and Explainability}}
\textcolor{black}{In this work, CFs are not only used for augmentation but also serve as interpretable explanations, providing insight into which features drive model predictions and how they can be modified to achieve a desired outcome.}
\begin{table*}[h]
    \small
    \centering
    
    \setlength{\tabcolsep}{2pt} 
    \caption{Example of LLM-suggested counterfactual intervention for a high-stress patient}
    \vspace{-1.5mm}
    {\renewcommand{\arraystretch}{1}
    \begin{tabular}{p{7in}}
        \toprule

        \multicolumn{1}{c}{\cellcolor{red!25}\textbf{Condition}} \\
        \midrule
        An 81-year-old patient labeled as “\textcolor{red}{stressed}” showed low deep sleep (30.1\%), moderate REM (15.4\%), high blood glucose (210.8 mg/dL), and limited activity (5.95 steps). Stress level was high (85.25), with poor glucose control (TIR: 12.5\%, 1 hyper event). \\[2mm]

        \midrule
        \multicolumn{1}{c}{\cellcolor{green!25}\textbf{Intervention}} \\

        \midrule
        The LLM suggests \textit{\textbf{increasing deep sleep}} to \textcolor{blue}{$\uparrow$}35\% and \textit{\textbf{REM sleep}} to \textcolor{blue}{$\uparrow$}20\%, which could help reduce physiological and emotional stress. It also recommends \textit{\textbf{lowering blood glucose}} from 210.8 to \textcolor{red}{$\downarrow$}180 mg/dL, aligning with better metabolic control. These changes reflect clinically actionable strategies such as sleep hygiene improvement and tighter glucose management. \\

        \bottomrule
    \end{tabular}
    }
    \label{tab:cf_intervention}
\end{table*}

\begin{table*}[h]
\small
\centering
\caption{Evaluating CFs on AI-READI Dataset for Class 0 and Class 1. ZS: Zero-shot, FS: Few-shot, \textbf{*}: Fine-tuned, \textcolor{red}{Red: best}, \textcolor{blue}{Blue}: second-best.}
{\renewcommand{\arraystretch}{1}

\vspace{-2mm}
    \begin{tabular}{
    p{0.7in}!{\vrule width 1pt}
    C{0.45in}C{0.45in}C{0.45in}C{0.6in}!{\vrule width 1pt}
    C{0.45in}C{0.45in}C{0.45in}C{0.6in}
    }
    \toprule
    \multirow{2}{*}{\textbf{Method}} 
    & \multicolumn{4}{c!{\vrule width 1pt}}{\textbf{Class 0 Metrics}} 
    & \multicolumn{4}{c}{\textbf{Class 1 Metrics}} \\
    \cmidrule{2-9}
    
    & \cellcolor{blue!18}\textbf{Validity $\uparrow$} 
    & \cellcolor{red!25}\textbf{Distance $\downarrow$}
    & \cellcolor{green!25}\textbf{Sparsity $\downarrow$}
    & \cellcolor{gray!18}\textbf{Plausibility $\uparrow$}
    & \cellcolor{blue!18}\textbf{Validity $\uparrow$} 
    & \cellcolor{red!25}\textbf{Distance $\downarrow$}
    & \cellcolor{green!25}\textbf{Sparsity $\downarrow$}
    & \cellcolor{gray!18}\textbf{Plausibility $\uparrow$} \\
    \midrule
    
    DiCE 
    & \cellcolor{blue!8}0.67 
    & \cellcolor{red!8}0.2 
    & \cellcolor{green!8}2.27 
    & \cellcolor{gray!8}\textcolor{red}{100}
    & \cellcolor{blue!8}0.58 
    & \cellcolor{red!8}0.41 
    & \cellcolor{green!8}2.4 
    & \cellcolor{gray!8}\textcolor{red}{99} \\
    
    CFNOW 
    & \cellcolor{blue!8}0.85 
    & \cellcolor{red!8}\textcolor{blue}{0.1} 
    & \cellcolor{green!8}2.9 
    & \cellcolor{gray!8}\textcolor{red}{100}
    & \cellcolor{blue!8}0.84 
    & \cellcolor{red!8}\textcolor{black}{0.25} 
    & \cellcolor{green!8}3 
    & \cellcolor{gray!8}\textcolor{red}{99} \\
    
    NICE 
    & \cellcolor{blue!8}0.44 
    & \cellcolor{red!8}\textcolor{red}{0.02} 
    & \cellcolor{green!8}\textcolor{red}{1.12} 
    & \cellcolor{gray!8}33
    & \cellcolor{blue!8}0.53 
    & \cellcolor{red!8}\textcolor{red}{0.04} 
    & \cellcolor{green!8}\textcolor{red}{1.31} 
    & \cellcolor{gray!8}35 \\
    
    \midrule
    
    GPT-4 (ZS) 
    & \cellcolor{blue!8}0.91 
    & \cellcolor{red!8}1.1 
    & \cellcolor{green!8}3.6 
    & \cellcolor{gray!8}85
    & \cellcolor{blue!8}0.89 
    & \cellcolor{red!8}1.5 
    & \cellcolor{green!8}3.8 
    & \cellcolor{gray!8}82 \\
    
    GPT-4 (FS) 
    & \cellcolor{blue!8}\textcolor{blue}{0.99} 
    & \cellcolor{red!8}1.2 
    & \cellcolor{green!8}4.4 
    & \cellcolor{gray!8}\textcolor{blue}{99}
    & \cellcolor{blue!8}\textcolor{blue}{0.92} 
    & \cellcolor{red!8}1.82 
    & \cellcolor{green!8}4 
    & \cellcolor{gray!8}\textcolor{blue}{96} \\
    
    BioMistral 
    & \cellcolor{blue!8}0.51 
    & \cellcolor{red!8}1.4 
    & \cellcolor{green!8}5.2 
    & \cellcolor{gray!8}77
    & \cellcolor{blue!8}0.47 
    & \cellcolor{red!8}1.5 
    & \cellcolor{green!8}4.1 
    & \cellcolor{gray!8}70 \\
    
    Llama 
    & \cellcolor{blue!8}0.62 
    & \cellcolor{red!8}1.6 
    & \cellcolor{green!8}4.6 
    & \cellcolor{gray!8}91
    & \cellcolor{blue!8}0.68 
    & \cellcolor{red!8}1.3 
    & \cellcolor{green!8}3.8 
    & \cellcolor{gray!8}78 \\
    
    \midrule
    
    BioMistral\textsuperscript{*}
    & \cellcolor{blue!8}\textcolor{red}{0.93} 
    & \cellcolor{red!8}0.92 
    & \cellcolor{green!8}\textcolor{black}{2.27} 
    & \cellcolor{gray!8}90
    & \cellcolor{blue!8}\textcolor{black}{0.91} 
    & \cellcolor{red!8}1.0 
    & \cellcolor{green!8}\textcolor{black}{2.1} 
    & \cellcolor{gray!8}\textcolor{black}{95} \\
    
    Llama\textsuperscript{*}
    & \cellcolor{blue!8}\textcolor{red}{0.99} 
    & \cellcolor{red!8}0.41
    & \cellcolor{green!8}\textcolor{blue}{1.8} 
    & \cellcolor{gray!8}\textcolor{red}{99}
    & \cellcolor{blue!8}\textcolor{red}{0.98} 
    & \cellcolor{red!8}\textcolor{blue}{0.2} 
    & \cellcolor{green!8}\textcolor{blue}{1.9} 
    & \cellcolor{gray!8}\textcolor{blue}{99} \\
    \bottomrule
    \end{tabular}
}
\label{tab:aireadi_result}
\end{table*}

The CF intervention in~\tblref{cf_intervention} illustrates how LLMs can propose clinically meaningful and physiologically grounded modifications for a high-stress patient. In this example, the model identifies low deep sleep, moderate REM sleep, elevated glucose (210.8 mg/dL), and low activity as major contributors to the stress prediction \textcolor{black}{showing how counterfactuals function as instance-level explanations, linking model predictions to clinically meaningful feature changes.}

~\tblref{aireadi_result} provides a comparison of counterfactual quality across baseline optimization methods, GPT-4, and open-source LLMs. While traditional methods such as DiCE, CFNOW, and NICE achieve strong plausibility and occasionally lower distances, they often propose unrealistic or non-actionable feature shifts. GPT-4 performs reliably in both zero-shot and few-shot modes, but its counterfactuals still exhibit larger feature deviations than desired for sensor-derived tabular data.

Fine-tuned BioMistral-7B and LLaMA-3.1-8B substantially improve validity, sparsity, and distance relative to their pretrained versions—showing gains of 20–40\% points in validity and reductions of more than 50\% in feature distance. Although fine-tuned LLMs do not always outperform every state-of-the-art baseline across all metrics, they remain highly competitive overall. Importantly, fine-tuned LLaMA provides the strongest balance across metrics, achieving near-perfect validity with minimal, clinically realistic modifications. 

Unlike optimization-based methods (e.g., DICE, CFNOW), which rely on access to model internals, our LLM-based approach generates CFs in a model-agnostic fashion while remaining interpretable and actionable. 

\begin{table*}[t]
\small
\centering
\caption{Performance Impact of LLM-Generated Counterfactuals Under Three Label-Scarcity Scenarios. 
Baseline performance is measured on the full training set. Negative values indicate the performance 
drop after undersampling. CF-added rows report the improvement relative to the reduced dataset.}
\label{tab:cf_scarcity_full_results}
\begin{tabular}{l l c c c c c}
\toprule
\textbf{Scenario} & \textbf{Method} & \textbf{ACC} & \textbf{PRE} & \textbf{REC} & \textbf{F1} & \textbf{AUC} \\
\midrule
Baseline (Using Full Training Set) & NN & 71.8 & 0.68 & 0.68 & 0.68 & 0.78 \\
\midrule
\multicolumn{7}{c}{\textbf{Scenario A — Positive-Class Undersampling (Class 1 Reduced by 50\%)}} \\
\midrule
Model trained on reduced data & NN & -12.87\%\textcolor{red}{$\downarrow$} & -11.76\%\textcolor{red}{$\downarrow$} & -16.18\%\textcolor{red}{$\downarrow$} & -14.71\%\textcolor{red}{$\downarrow$} & -14.10\%\textcolor{red}{$\downarrow$} \\
\cmidrule(lr){2-7}
\multirow{5}{*}{CF Added (Recovery over Reduced)} 
& BioMistral & 7.10\%\textcolor{blue}{$\uparrow$} & 5.00\%\textcolor{blue}{$\uparrow$} & 8.77\%\textcolor{blue}{$\uparrow$} & 6.90\%\textcolor{blue}{$\uparrow$} & 7.46\%\textcolor{blue}{$\uparrow$} \\
& LLaMA & 8.54\%\textcolor{blue}{$\uparrow$} & 6.67\%\textcolor{blue}{$\uparrow$} & 10.53\%\textcolor{blue}{$\uparrow$} & 8.62\%\textcolor{blue}{$\uparrow$} & 7.46\%\textcolor{blue}{$\uparrow$} \\
& GPT-4 & 10.29\%\textcolor{blue}{$\uparrow$} & 10.00\%\textcolor{blue}{$\uparrow$} & 12.28\%\textcolor{blue}{$\uparrow$} & 12.07\%\textcolor{blue}{$\uparrow$} & 11.94\%\textcolor{blue}{$\uparrow$} \\
& BioMistral\textsuperscript{*} & 14.93\%\textcolor{blue}{$\uparrow$} & 18.33\%\textcolor{blue}{$\uparrow$} & 22.81\%\textcolor{blue}{$\uparrow$} & 20.69\%\textcolor{blue}{$\uparrow$} & 20.90\%\textcolor{blue}{$\uparrow$} \\
& LLaMA\textsuperscript{*} & 21.00\%\textcolor{blue}{$\uparrow$} & 20.00\%\textcolor{blue}{$\uparrow$} & 24.56\%\textcolor{blue}{$\uparrow$} & 22.41\%\textcolor{blue}{$\uparrow$} & 25.37\%\textcolor{blue}{$\uparrow$} \\
\midrule
\multicolumn{7}{c}{\textbf{Scenario B — Negative-Class Undersampling (Class 0 Reduced by 50\%)}} \\
\midrule
Model trained on reduced data & NN & -10.72\%\textcolor{red}{$\downarrow$} & -8.82\%\textcolor{red}{$\downarrow$} & -10.29\%\textcolor{red}{$\downarrow$} & -10.29\%\textcolor{red}{$\downarrow$} & -12.82\%\textcolor{red}{$\downarrow$} \\
\cmidrule(lr){2-7}
\multirow{5}{*}{CF Added (Recovery over Reduced)} 
& BioMistral & 4.52\%\textcolor{blue}{$\uparrow$} & 1.61\%\textcolor{blue}{$\uparrow$} & 1.64\%\textcolor{blue}{$\uparrow$} & 1.64\%\textcolor{blue}{$\uparrow$} & 4.41\%\textcolor{blue}{$\uparrow$} \\
& LLaMA & 5.93\%\textcolor{blue}{$\uparrow$} & 3.23\%\textcolor{blue}{$\uparrow$} & 3.28\%\textcolor{blue}{$\uparrow$} & 3.28\%\textcolor{blue}{$\uparrow$} & 5.88\%\textcolor{blue}{$\uparrow$} \\
& GPT-4 & 6.86\%\textcolor{blue}{$\uparrow$} & 4.84\%\textcolor{blue}{$\uparrow$} & 4.92\%\textcolor{blue}{$\uparrow$} & 4.92\%\textcolor{blue}{$\uparrow$} & 8.82\%\textcolor{blue}{$\uparrow$} \\
& BioMistral\textsuperscript{*} & 14.82\%\textcolor{blue}{$\uparrow$} & 14.52\%\textcolor{blue}{$\uparrow$} & 13.11\%\textcolor{blue}{$\uparrow$} & 14.75\%\textcolor{blue}{$\uparrow$} & 20.59\%\textcolor{blue}{$\uparrow$} \\
& LLaMA\textsuperscript{*} & 17.16\%\textcolor{blue}{$\uparrow$} & 16.13\%\textcolor{blue}{$\uparrow$} & 16.39\%\textcolor{blue}{$\uparrow$} & 16.39\%\textcolor{blue}{$\uparrow$} & 23.53\%\textcolor{blue}{$\uparrow$} \\
\midrule
\multicolumn{7}{c}{\textbf{Scenario C — Dual-Class Undersampling (Both Classes Reduced by 50\%)}} \\
\midrule
Model trained on reduced data & NN & -11.14\%\textcolor{red}{$\downarrow$} & -10.29\%\textcolor{red}{$\downarrow$} & -13.24\%\textcolor{red}{$\downarrow$} & -11.76\%\textcolor{red}{$\downarrow$} & -14.10\%\textcolor{red}{$\downarrow$} \\
\cmidrule(lr){2-7}
\multirow{5}{*}{CF Added (Recovery over Reduced)} 
& BioMistral & 4.55\%\textcolor{blue}{$\uparrow$} & 1.64\%\textcolor{blue}{$\uparrow$} & 3.39\%\textcolor{blue}{$\uparrow$} & 1.67\%\textcolor{blue}{$\uparrow$} & 4.48\%\textcolor{blue}{$\uparrow$} \\
& LLaMA & 5.90\%\textcolor{blue}{$\uparrow$} & 4.92\%\textcolor{blue}{$\uparrow$} & 6.78\%\textcolor{blue}{$\uparrow$} & 5.00\%\textcolor{blue}{$\uparrow$} & 7.46\%\textcolor{blue}{$\uparrow$} \\
& GPT-4 & 6.27\%\textcolor{blue}{$\uparrow$} & 4.92\%\textcolor{blue}{$\uparrow$} & 5.08\%\textcolor{blue}{$\uparrow$} & 5.00\%\textcolor{blue}{$\uparrow$} & 8.96\%\textcolor{blue}{$\uparrow$} \\
& BioMistral\textsuperscript{*}& 15.99\%\textcolor{blue}{$\uparrow$} & 16.39\%\textcolor{blue}{$\uparrow$} & 16.95\%\textcolor{blue}{$\uparrow$} & 16.67\%\textcolor{blue}{$\uparrow$} & 22.39\%\textcolor{blue}{$\uparrow$} \\
& LLaMA\textsuperscript{*} & 21.47\%\textcolor{blue}{$\uparrow$} & 19.67\%\textcolor{blue}{$\uparrow$} & 22.03\%\textcolor{blue}{$\uparrow$} & 20.00\%\textcolor{blue}{$\uparrow$} & 26.87\%\textcolor{blue}{$\uparrow$} \\
\bottomrule
\end{tabular}
\end{table*}

\begin{figure}[!h]
\centering
\includegraphics[width=0.9\linewidth, trim={265 175 250 65},clip]{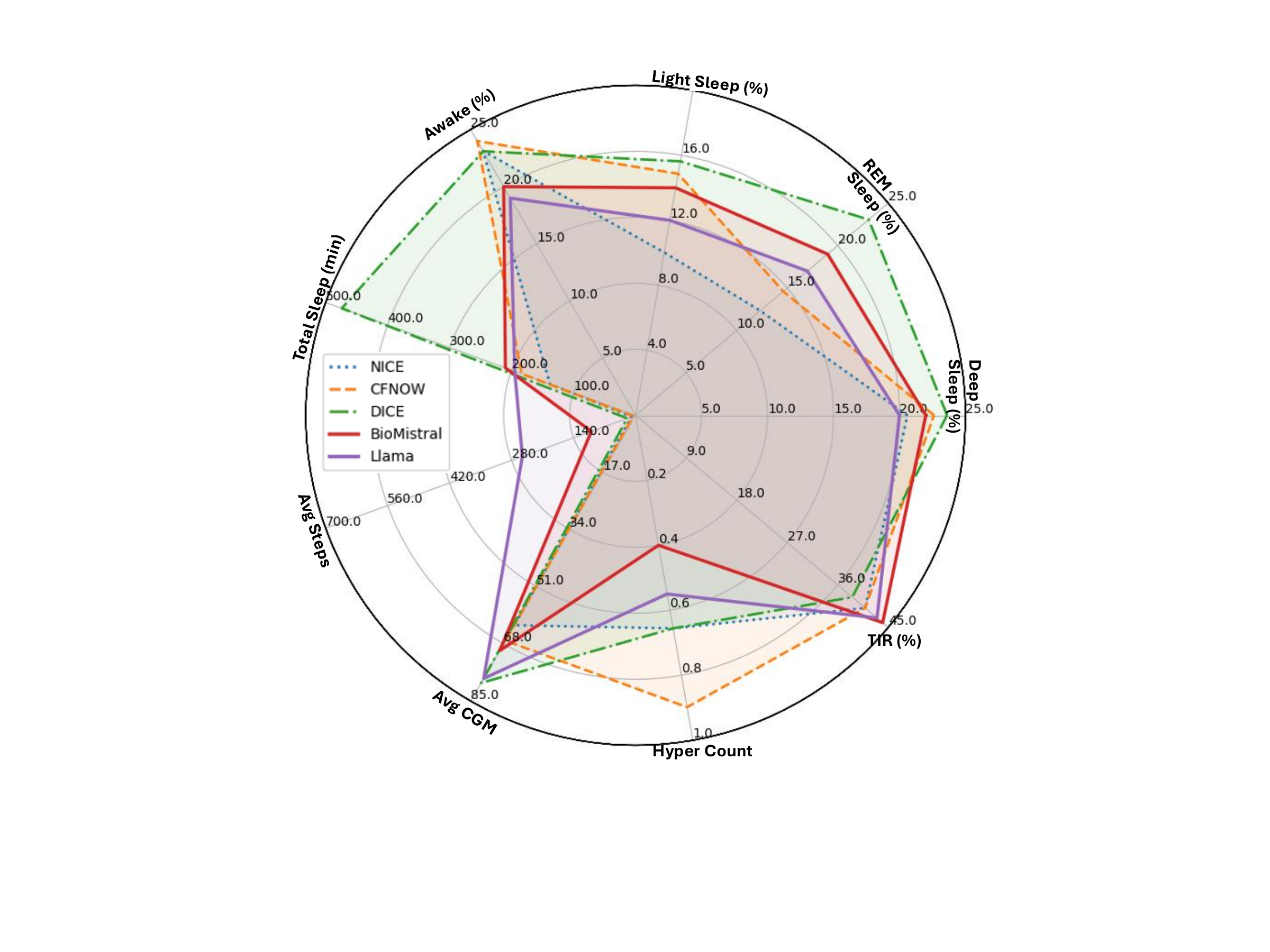}
\caption{Feature diversity in the generated CFs for AI-Readi data. Avg: Average, Hyper: No. of hyperglycemia}
\label{fig:diversity}
\vspace{-2mm}
\end{figure}
\textbf{Feature Diversity}: \textcolor{black}{We define feature diversity as the distribution and magnitude of feature modifications across counterfactual samples, indicating how different methods explore the feature space. In this context, diversity is desirable when it occurs over actionable features, as it reflects meaningful intervention strategies, but less useful when concentrated on non-modifiable physiological attributes.} The radar plot (\figref{diversity}) highlights important differences in actionability and realism across counterfactual generation methods. Traditional CF methods such as DICE and CFNOW often propose large shifts in physiological features- specially sleep-stage percentages and total sleep duration that are not immediately modifiable in real-life settings and therefore have lower practical utility. NICE performs better by recommending smaller, more localized changes, but it still occasionally alters features like REM or deep sleep that individuals cannot directly control in the short term. In contrast, the fine-tuned BioMistral and Llama models produce counterfactuals concentrated around \textit{highly actionable variables}, such as average steps, glucose levels (Average CGM, TIR\%), and hyperglycemia frequency—factors that can be modified through short-term behavioral or treatment adjustments. This shift toward modifiable lifestyle and metabolic variables makes the LLM-based counterfactuals more clinically realistic and better aligned with interventions that individuals can adopt immediately. Overall, the Llama model generates the most actionable and plausible counterfactual recommendations, avoiding unrealistic alterations to intrinsic sleep architecture while still achieving high validity in flipping the prediction.

\subsection{Augmentation}
\begin{figure*}[!h]
\centering
\includegraphics[width=1\linewidth, trim={0 260 0 310},clip]{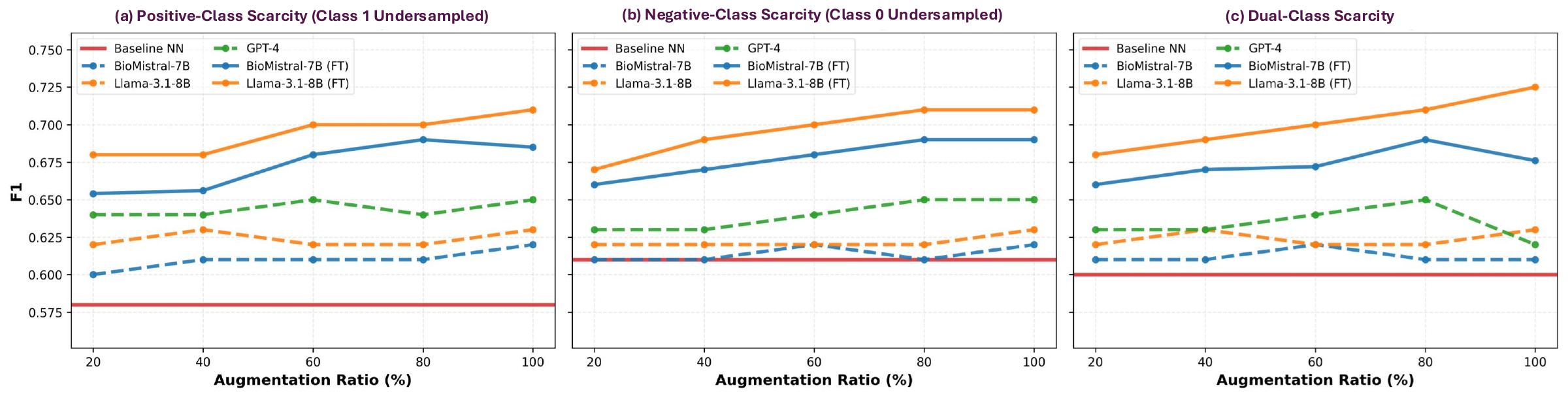}
\caption{Impact of LLM-Generated Counterfactual Augmentation Under Class-Specific Label Scarcity}
\label{fig:cf-effect}
\vspace{-4mm}
\end{figure*}

\subsubsection{Performance Recovery}
Table~\ref{tab:cf_scarcity_full_results} summarizes the impact of LLM-generated counterfactual samples on classifier performance under three controlled label-scarcity settings. We first trained a neural network on the full training set to establish a baseline. Next, we constructed three undersampling scenarios: (1) Positive-Class Scarcity (Class 1 undersampled), (2) Negative-Class Scarcity (Class 0 undersampled), and (3) Dual-Class Scarcity (both classes undersampled). In each case, removing 50\% of the samples from the targeted class resulted in a substantial performance drop across all metrics, as reflected by negative deltas relative to the baseline neural network. To recover the lost information, we generated counterfactual samples using several LLMs (BioMistral-7B, LLaMA-3.1-8B, and GPT-4o) in both zero-shot and fine-tuned configurations, and added only the counterfactuals corresponding to the class that was undersampled. The ``CF Added'' rows report the performance recovery relative to the reduced-data model, showing that fine-tuned LLMs consistently provide the largest improvement, with fine-tuned LLaMA delivering the strongest gains across all three scarcity scenarios.

LLaMA-3.1-8B (FT) restores more than 22\% of F1 in the positive-class scarcity setting and roughly 20\% under dual-class scarcity—fully reversing the degradation introduced by undersampling. Even in the more challenging negative-class scarcity case, fine-tuned LLMs recover 16–17\% of F1, clearly outperforming their zero-shot counterparts. These large percentage-point gains highlight the strong corrective effect of CF-based augmentation when guided by fine-tuned LLMs.

\textcolor{black}{To validate model-agnostic behavior, we evaluate CF augmentation across multiple classifiers (NN, RF, SVC, XGB). As shown in~\tblref{cf_augmentation_performance} in the supplementary~\secref{sup_model_agnostic}, performance improvements are consistent across all models and scenarios, demonstrating that the proposed framework generalizes well and is not tied to a specific classifier.\newline
We also evaluated prompt sensitivity using alternative prompt formulations and observed only minor variations in F1 scores, suggesting that model performance is largely driven by LLM capability and fine-tuning rather than prompt design (see supplementary \secref{sup_prompt_sen}).
}

\subsubsection{Effect of Augmentation Ratio} ~\figref{cf-effect} further explores how the amount of counterfactual augmentation influences performance by varying the augmentation ratio from 20\% to 100\%. Each subplot begins with the reduced-data model in scenario A, B, or C, where the neural network exhibits markedly degraded performance due to undersampling; this is shown as the horizontal dashed baseline in~\figref{cf-effect} (A-C). Starting from these weakened models, we incrementally add different quantities of LLM-generated counterfactuals and observe how performance recovers as augmentation increases. Across all three scarcity scenarios, the F1 score improves steadily as more counterfactuals are injected into the training set. The effect is most pronounced for the fine-tuned LLaMA-3.1-8B and BioMistral-7B models, which show clear monotonic gains and achieve the highest F1 scores at large augmentation ratios. 

Zero-shot LLMs (dashed lines) still provide measurable improvement over the undersampled baseline, but their gains plateau earlier, suggesting limited ability to mimic the true class manifold without task-specific adaptation. Another consistent observation is that positive-class scarcity (Scenario A) and dual-class scarcity (Scenario C) benefit more strongly from CF augmentation than negative-class scarcity (Scenario B), indicating that the model is particularly sensitive to missing minority-class examples. Overall, the F1 curves demonstrate that counterfactuals function as an effective and controllable data augmenter, with augmentation magnitude and LLM fine-tuning both playing significant roles in performance recovery.

\textcolor{black}{Beyond performance trends, an important consideration is whether generated CFs remain on the underlying data manifold as augmentation increases. Our results suggest that this assumption largely holds. The high plausibility scores (\tblref{aireadi_result}) indicate that CFs lie within the empirical feature distribution, while the latent-space visualization (Fig. 8) shows that fine-tuned LLaMA generates CFs that align closely with the target class clusters rather than forming out-of-distribution samples. Additionally,~\figref{cf-effect} demonstrates stable and monotonic performance improvements across augmentation ratios, with no evidence of degradation at higher levels. However, we observe that performance gains begin to plateau beyond approximately 60–80\% augmentation, suggesting that moderate augmentation ratios may be sufficient in practice to balance diversity and realism.}
\begin{figure}[!htb]
\vspace{-3mm}
\centering
\includegraphics[width=0.8\linewidth, trim={381 250 370 75},clip]{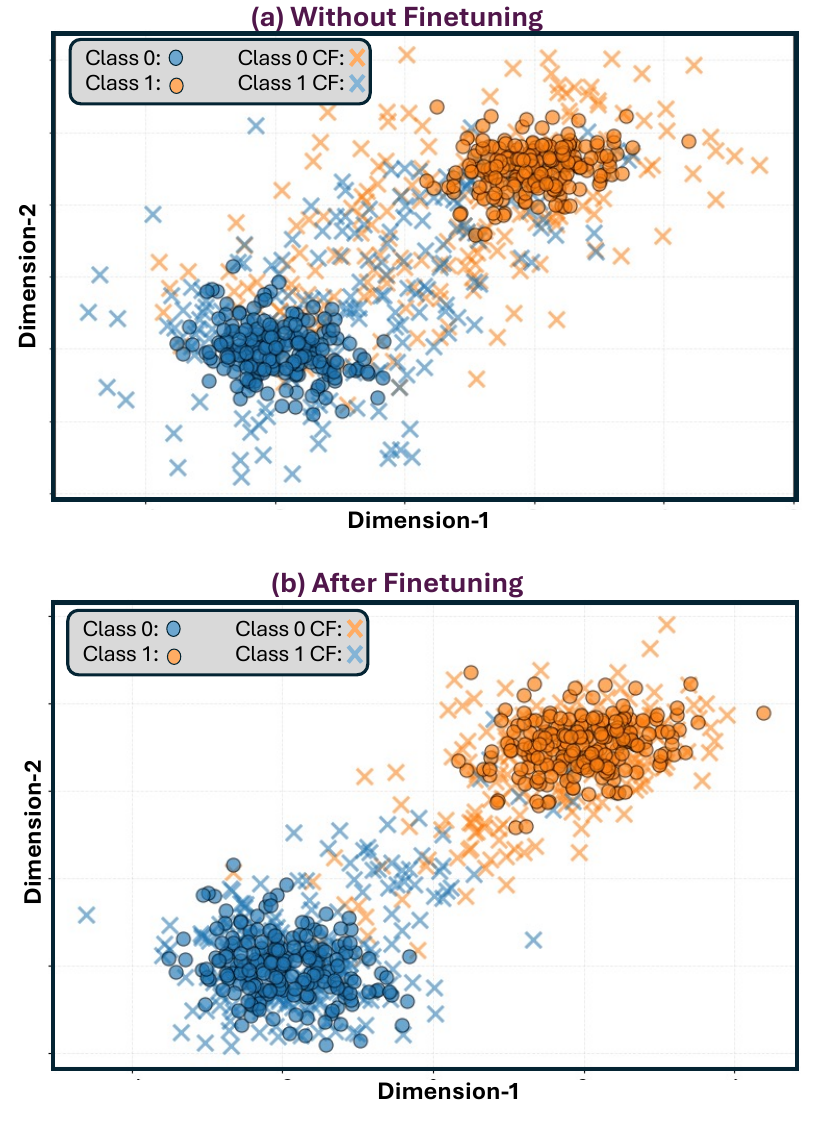}
\caption{Latent Space Distribution of Factual and Counterfactual Samples Generated by LLaMA-3.1-8B on the AI-READI Dataset (Before vs. After Fine-Tuning). The amount of overlap between two classes is much lesser in the dataset augmented using fine-Tuned LLM.}
\label{fig:umap}
\vspace{-4mm}
\end{figure}
\subsection{Ablation Study} 
Fine-tuning LLaMA-3.1-8B leads to markedly improved counterfactual behavior, as shown in~\figref{umap}. Using a randomly selected subset of 200 factual samples, the zero-shot model generates diffused and often misaligned CFs; the fine-tuned model produces CFs that cluster tightly within the appropriate class region and maintain local structural coherence. This improved spatial organization highlights the effectiveness of finetuning in producing plausible, semantically meaningful, and distribution-consistent counterfactuals.

\section{\textbf{DISCUSSION} }
This study presents the first systematic evaluation of LLM-generated counterfactuals for intervention design and data augmentation in sensor-derived health datasets. We evaluate multiple LLM families—GPT-4o, BioMistral-7B, and LLaMA-3.1-8B—in both pretrained and fine-tuned configurations, demonstrating that fine-tuned LLaMA achieves an average 19.9\% F1 recovery across three scarcity conditions when 50\% of labels are removed. This represents a substantial improvement in classifier robustness under label scarcity while providing interpretable, domain-aligned explanations that traditional CF methods cannot offer. \textcolor{black}{Since the framework does not depend on model internals, it is inherently robust to the rapid evolution of LLMs, and newer models can be incorporated in a plug-and-play manner to further improve performance.}
\textcolor{black}{We follow standard counterfactual evaluation practice by reporting validity, distance/proximity, sparsity, and plausibility as the primary CF-quality metrics. Additional criteria such as robustness to perturbations, noise sensitivity, and personalized feasibility constraints are important but depend on the downstream classifier, user-specific restrictions, or human-in-the-loop feedback; therefore, we treat them as future extensions rather than primary metrics in this static post-hoc setting.}

Our comparative analysis reveals a critical distinction between LLM-based and optimization-based approaches: while traditional methods (DiCE, CFNOW, NICE) frequently modify clinically intractable features such as sleep architecture, fine-tuned LLMs prioritize actionable variables—average steps, glucose metrics (Avg CGM, TIR\%), and hyperglycemia frequency—that reflect realistic behavioral interventions. Latent-space visualizations confirm that fine-tuned LLaMA produces semantically coherent CFs aligned with the target class manifold, addressing a key limitation of existing methods.

\subsection{Clinical Implications}
These findings highlight the potential for LLM-generated counterfactuals to support personalized health guidance. By suggesting small, clinically plausible adjustments grounded in sensor-derived physiology, CFs can help identify early behavioral or metabolic interventions for individuals at risk. Moreover, CF-based augmentation offers a scalable alternative to collecting more sensor data, helping improve classifier robustness in health settings where labeled data are scarce.
\subsection{Dataset \& Limitation}
We selected the AI-READI dataset because publicly available clinical datasets suitable for LLM fine-tuning on structured tabular health features are extremely limited. Most open-source biomedical datasets are too small, lack multimodal coverage, or contain insufficient feature richness for effective LLM alignment. AI-READI, by contrast, offers a large, well-curated combination of CGM, activity, sleep, demographic, and stress features—making it one of the few datasets capable of supporting robust LLM fine-tuning and CF evaluation.

Despite strong overall performance, fine-tuned LLMs do not outperform all state-of-the-art CF baselines across every metric, particularly in distance or sparsity for certain classes. Some CFs may still exceed clinically achievable ranges, and fine-tuning requires labeled data that may be scarce for some conditions. \textcolor{black}{Because CFs are post-hoc explanations of a trained predictor, any bias or error in the underlying classifier may propagate to the generated counterfactuals; therefore, fairness-aware classifier training and bias-sensitive CF evaluation remain important future directions.}
Finally, our work is limited to structured tabular features and does not yet incorporate multimodal raw sensor streams or clinical text. 

\subsection{Future Directions}
Several promising avenues remain for improving LLM-based counterfactual reasoning. First, integrating LLMs directly into the machine learning training loop could enable iterative, CF-guided model updates (“LLM-in-the-loop learning”). Second, incorporating clinical knowledge graphs or causal structures into the fine-tuning pipeline may reduce unrealistic or physiologically implausible feature changes. \textcolor{black}{Third, future work should incorporate physician-in-the-loop evaluation and expert validation to assess the clinical utility, safety, and interpretability of generated counterfactual interventions.} Expanding beyond tabular data to multimodal LLMs could further enable CF generation from raw sensor traces, free-text clinical notes, or imaging.
Lastly, deploying CF-driven augmentation in real-world digital health ecosystems may help assess the longer-term impact of CF-based guidance on early intervention and patient outcomes.

\section{\textbf{CONCLUSION}}
In this work, we introduce a novel framework for generating CFs using large language models (LLMs) and extend it to evaluate both proprietary and open-source models across multiple generative settings. Our results show that LLM-generated counterfactuals are semantically coherent, clinically plausible, and capable of improving downstream robustness when used for data augmentation—restoring, on average, ~20\% F1 under severe label scarcity. Fine-tuned LLaMA and BioMistral models, in particular, produce compact and actionable CFs that outperform their pretrained counterparts and remain competitive with state-of-the-art optimization methods.

\textcolor{black}{To the best of our knowledge, this is among the first studies to evaluate fine-tuned LLMs as structured counterfactual generators for both explanation and augmentation in sensor-driven health settings.}

\section*{Acknowledgment}
This work was supported in part by the National Science Foundation (NSF) under Grant IIS-2402650. 
A. Arefeen was supported in part by the National Institute of Diabetes and Digestive and Kidney Diseases (NIDDK) of the National Institutes of Health (NIH) under Award T32DK137525. 
The content is solely the responsibility of the authors and does not necessarily represent the official views of the NSF or NIH.

\section*{Supplementary Materials}
\subsection{\color{black}Model-Agnostic Evaluation}
\label{sec:sup_model_agnostic}
\textcolor{black}{
To further validate the model-agnostic nature of the proposed framework, we evaluate the effectiveness of CF-based augmentation across multiple downstream classifiers. Specifically, we use the same set of counterfactual samples generated by the fine-tuned LLaMA-3.1-8B model and apply them consistently across three label-scarcity scenarios: Positive-Class Scarcity (Scene A), Negative-Class Scarcity (Scene B), and Dual-Class Scarcity (Scene C). We then train four different classifiers—Neural Network (NN), Random Forest (RF), Support Vector Machine (SVC), and XGBoost (XGB)—on both the reduced and CF-augmented datasets.
}
\textcolor{black}{
As shown in~\tblref{cf_augmentation_performance}, CF augmentation leads to consistent performance improvements across all classifiers and scenarios, with F1 scores converging to a similar range regardless of the underlying model. The variation across classifiers remains minimal, indicating that the benefits of counterfactual augmentation are not dependent on a specific classifier architecture. These findings provide strong empirical evidence that the proposed framework is robust and generalizable, supporting its applicability in diverse machine learning settings.
}

\begin{table}[ht]
\color{black}
\centering
\caption{\textcolor{black}{Model-agnostic performance of CF augmentation across classifiers. Values in () represent non-augmented results. }}
\vspace{-1.5mm}
\setlength{\tabcolsep}{2.7pt} 

    \renewcommand{\arraystretch}{1}
\begin{tabular}{lcccc}
\toprule
\textbf{Classifier} & \textbf{NN} & \textbf{RF} & \textbf{SVC} & \textbf{XGB} \\
\hline
Full dataset & 0.68 & 0.69 & 0.64 & 0.69 \\
\midrule
Scene A & 0.83 (0.58) & 0.80 (0.60) & 0.79 (0.60) & 0.82 (0.62) \\
Scene B & 0.79 (0.61) & 0.78 (0.63) & 0.77 (0.61) & 0.79 (0.60) \\
Scene C & 0.82 (0.60) & 0.81 (0.62) & 0.81 (0.59) & 0.82 (0.59) \\
\bottomrule
\end{tabular}
\label{tab:cf_augmentation_performance}
\end{table}

\subsection{\color{black}Prompt Sensitivity Analysis}
\label{sec:sup_prompt_sen}
\begin{figure}[!h]
\centering
\includegraphics[width=1\linewidth, trim={3 9 4 6},clip]{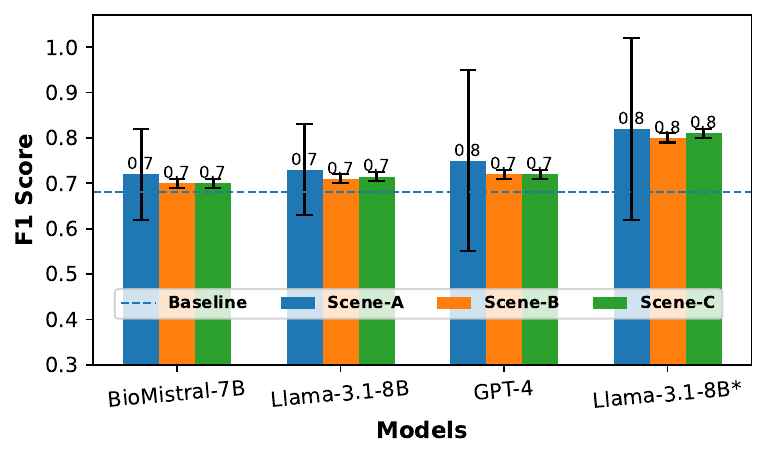}
\caption{\color{black}Impact of prompt variations on F1 score across LLMs.}
\label{fig:prmopt_f1}
\vspace{-2mm}
\end{figure}
\textcolor{black}{
To assess the sensitivity of our framework to prompt design, we evaluated two alternative prompt formulations: (i) a minimality-focused prompt emphasizing sparse feature changes, and (ii) an actionability-focused prompt prioritizing clinically modifiable variables. These experiments were conducted using the fine-tuned LLaMA-3.1-8B model across all three label-scarcity scenarios. As summarized in~\figref{prmopt_f1}, the variations in performance across prompt types are relatively small, with F1 scores remaining stable across settings. 
\textcolor{black}{Results are reported as mean ± standard deviation across repeated evaluations, where the observed variance remains small ($\sigma \leq$0.08), indicating stable augmentation behavior across prompt variations.}
While the minimality-focused prompt tends to produce more compact counterfactuals and the actionability-focused prompt shifts feature selection toward behaviorally relevant attributes, these differences do not significantly impact downstream classification performance. This suggests that the overall effectiveness of the proposed framework is primarily driven by the underlying LLM capability and fine-tuning, rather than prompt design, indicating robustness to reasonable prompt variations.
}

\subsection{Dataset Diversity}
\begin{figure}[!htb]
\vspace{-2mm}
\centering
\includegraphics[width=1\linewidth,trim={4 4 4 20},clip]{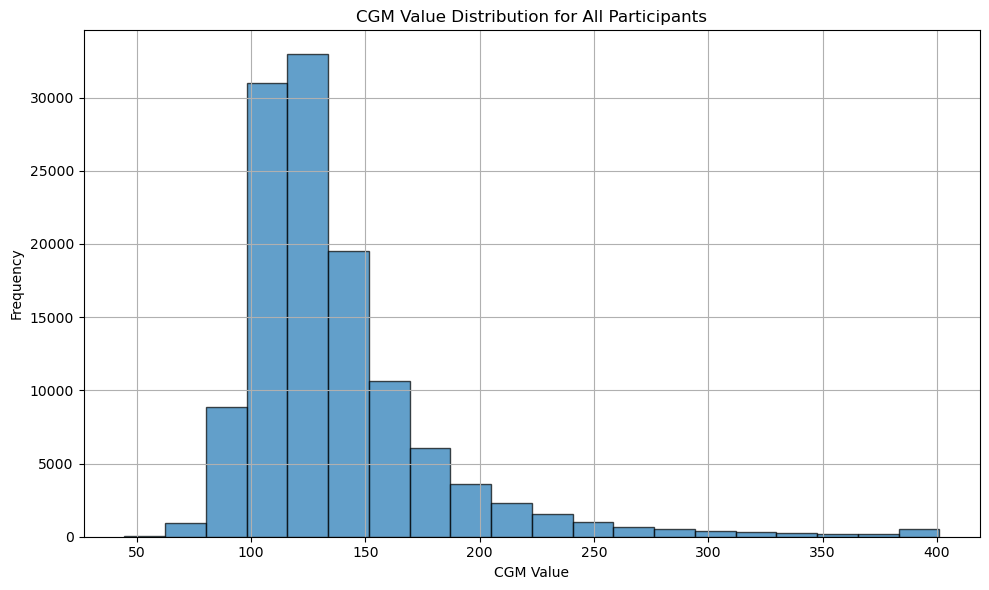}
\caption{Distribution of CGM Values Across AI-READI Participants}
\label{fig:cgm_distribution}
\vspace{-2mm}
\end{figure}
\begin{figure}[!htb]
\vspace{-2mm}
\centering
\includegraphics[width=1\linewidth,trim={4 4 4 20},clip]{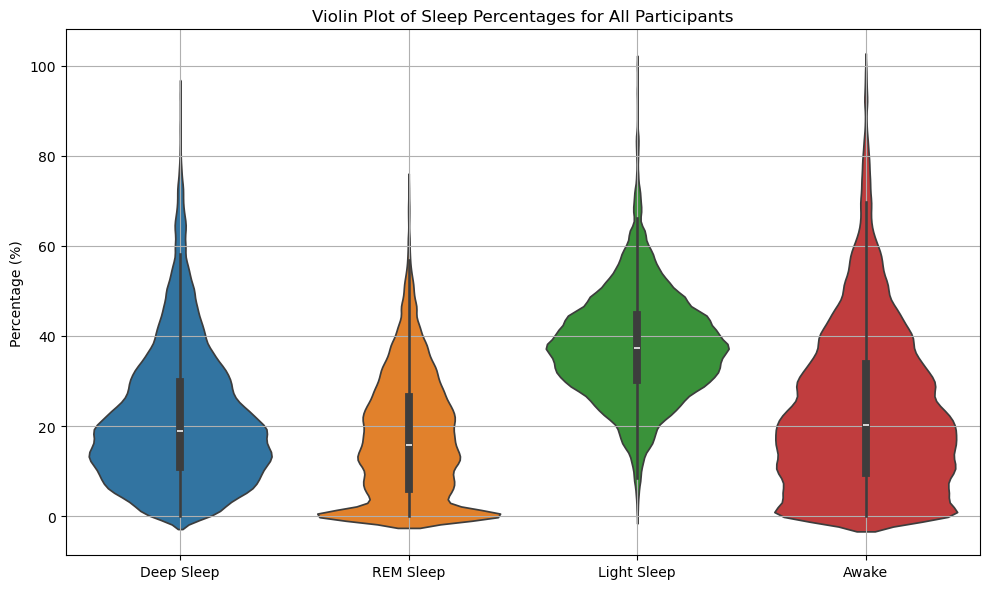}
\caption{Distribution of Sleep-Stage Percentages Across AI-READI Participants}
\label{fig:sleep_distribution}
\vspace{-4mm}
\end{figure}
\textcolor{black}{
To further characterize the diversity of the AI-READI dataset, \figref{cgm_distribution} and \figref{sleep_distribution} visualize the distributions of metabolic and physiological features across participants. The CGM histogram demonstrates a broad range of glucose values spanning normoglycemic, pre-diabetic, and hyperglycemic conditions, reflecting substantial metabolic heterogeneity across the population. Similarly, the sleep-stage violin plots show high variability in deep sleep, REM sleep, light sleep, and awake percentages, indicating diverse behavioral and physiological patterns.
}
\begin{table}[h]
\centering
\caption{\color{black}Cohort-wise augmentation performance (F1 score) across different diabetes-management groups for Scene-C. Fine-tuned LLMs consistently outperform their pretrained counterparts across all cohorts. * Fine-tuned model}
\label{tab:azt1d_cohort_result}
\setlength{\tabcolsep}{3pt} 
\renewcommand{\arraystretch}{0.7}
\color{black}
\begin{tabular}{lcccc}
\toprule
\textbf{Cohort} & \textbf{BioMistral} & \textbf{BioMistral*} & \textbf{LLaMA} & \textbf{LLaMA*} \\
\midrule
Healthy    & 0.71 & 0.80 & 0.73 & \textbf{0.84} \\
Pre-T2DM & 0.69 & 0.79 & 0.72 & \textbf{0.83} \\
Oral  & 0.67 & 0.77 & 0.70 & \textbf{0.81} \\
Insulin & 0.94 & 0.75 & 0.68 & \textbf{0.79} \\
\bottomrule
\end{tabular}
\end{table}
\textcolor{black}{
We additionally observed consistent augmentation trends across these cohorts, with fine-tuned LLMs outperforming pretrained models in all groups (shown in \tblref{azt1d_cohort_result}. Although insulin-controlled participants remained more challenging due to higher metabolic variability, fine-tuned LLaMA maintained stable performance across cohorts, suggesting that the framework generalizes across diverse disease-management conditions rather than a single homogeneous population.}

\bibliographystyle{IEEEtran}
\bibliography{ref.bib}

\begin{thebibliography}{10}
\providecommand{\url}[1]{#1}
\csname url@samestyle\endcsname
\providecommand{\newblock}{\relax}
\providecommand{\bibinfo}[2]{#2}
\providecommand{\BIBentrySTDinterwordspacing}{\spaceskip=0pt\relax}
\providecommand{\BIBentryALTinterwordstretchfactor}{4}
\providecommand{\BIBentryALTinterwordspacing}{\spaceskip=\fontdimen2\font plus
\BIBentryALTinterwordstretchfactor\fontdimen3\font minus \fontdimen4\font\relax}
\providecommand{\BIBforeignlanguage}[2]{{%
\expandafter\ifx\csname l@#1\endcsname\relax
\typeout{** WARNING: IEEEtran.bst: No hyphenation pattern has been}%
\typeout{** loaded for the language `#1'. Using the pattern for}%
\typeout{** the default language instead.}%
\else
\language=\csname l@#1\endcsname
\fi
#2}}
\providecommand{\BIBdecl}{\relax}
\BIBdecl

\bibitem{soumma2025ai}
S.~B. Soumma, A.~Mamun, and H.~Ghasemzadeh, ``Ai-powered wearable sensors for health monitoring and clinical decision making,'' \emph{Current Opinion in Biomedical Engineering}, p. 100628, 2025.

\bibitem{Wachter2017CounterfactualEW}
S.~Wachter, B.~D. Mittelstadt, and C.~Russell, ``Counterfactual explanations without opening the black box: Automated decisions and the gdpr,'' \emph{Cybersecurity}, 2017.

\bibitem{Guidotti2018ASO}
R.~Guidotti, A.~Monreale, F.~Turini, D.~Pedreschi, and F.~Giannotti, ``A survey of methods for explaining black box models,'' \emph{ACM Computing Surveys (CSUR)}, vol.~51, pp. 1 -- 42, 2018.

\bibitem{Karimi2020AlgorithmicRF}
A.-H. Karimi, B.~Scholkopf, and I.~Valera, ``Algorithmic recourse: from counterfactual explanations to interventions,'' \emph{Proceedings of the 2021 ACM Conference on Fairness, Accountability, and Transparency}, 2020.

\bibitem{Karimi2019ModelAgnosticCE}
A.-H. Karimi, G.~Barthe, B.~Balle, and I.~Valera, ``Model-agnostic counterfactual explanations for consequential decisions,'' \emph{ArXiv}, vol. abs/1905.11190, 2019.

\bibitem{mothilal2020dice}
R.~K. Mothilal, A.~Sharma, and C.~Tan, ``Explaining machine learning classifiers through diverse counterfactual explanations,'' in \emph{Proceedings of the 2020 Conference on Fairness, Accountability, and Transparency}, 2020, pp. 607--617.

\bibitem{DEOLIVEIRA2023}
R.~M.~B. {de Oliveira}, K.~Sörensen, and D.~Martens, ``A model-agnostic and data-independent tabu search algorithm to generate counterfactuals for tabular, image, and text data,'' \emph{European Journal of Operational Research}, 2023.

\bibitem{Brughmans2021NICEAA}
D.~Brughmans and D.~Martens, ``Nice: an algorithm for nearest instance counterfactual explanations,'' \emph{Data Mining and Knowledge Discovery}, pp. 1--39, 2021.

\bibitem{mann2020language}
B.~Mann, N.~Ryder, M.~Subbiah, J.~Kaplan, P.~Dhariwal, A.~Neelakantan, P.~Shyam, G.~Sastry, A.~Askell, S.~Agarwal \emph{et~al.}, ``Language models are few-shot learners,'' \emph{arXiv preprint arXiv:2005.14165}, vol.~1, p.~3, 2020.

\bibitem{soumma2025sensecf}
\BIBentryALTinterwordspacing
S.~B. Soumma, A.~Arefeen, S.~M. Carpenter, M.~Hingle, and H.~Ghasemzadeh, ``Sense{CF}: {LLM}-prompted counterfactuals for intervention and sensor data augmentation,'' in \emph{IEEE-EMBS International Conference on Body Sensor Networks 2025}, 2025. [Online]. Available: \url{https://openreview.net/forum?id=8qqMeF9EmT}
\BIBentrySTDinterwordspacing

\bibitem{fizle_huan}
\BIBentryALTinterwordspacing
A.~Bhattacharjee, R.~Moraffah, J.~Garland, and H.~Liu, ``{ Zero-shot LLM-guided Counterfactual Generation: A Case Study on NLP Model Evaluation },'' in \emph{2024 IEEE International Conference on Big Data (BigData)}.\hskip 1em plus 0.5em minus 0.4em\relax Los Alamitos, CA, USA: IEEE Computer Society, Dec. 2024, pp. 1243--1248. [Online]. Available: \url{https://doi.ieeecomputersociety.org/10.1109/BigData62323.2024.10825537}
\BIBentrySTDinterwordspacing

\bibitem{li-etal-2024-prompting}
\BIBentryALTinterwordspacing
Y.~Li, M.~Xu, X.~Miao, S.~Zhou, and T.~Qian, ``Prompting large language models for counterfactual generation: An empirical study,'' in \emph{Proceedings of the 2024 Joint International Conference on Computational Linguistics, Language Resources and Evaluation (LREC-COLING 2024)}, N.~Calzolari, M.-Y. Kan, V.~Hoste, A.~Lenci, S.~Sakti, and N.~Xue, Eds.\hskip 1em plus 0.5em minus 0.4em\relax Torino, Italia: ELRA and ICCL, May 2024, pp. 13\,201--13\,221. [Online]. Available: \url{https://aclanthology.org/2024.lrec-main.1156/}
\BIBentrySTDinterwordspacing

\bibitem{Chen2025CounterBenchAB}
Y.~Chen, V.~K. Singh, J.~Ma, and R.~Tang, ``Counterbench: A benchmark for counterfactuals reasoning in large language models,'' \emph{ArXiv}, vol. abs/2502.11008, 2025.

\bibitem{Russell2019EfficientSF}
C.~Russell, ``Efficient search for diverse coherent explanations,'' \emph{Proceedings of the Conference on Fairness, Accountability, and Transparency}, 2019.

\bibitem{ai2024ai}
P.~M. http://orcid. org/0000-0001-6343-2140 Drolet Caroline 4 http://orcid. org/0000-0003-2287-4190 Lucero Abigail 8 Matthies Dawn 7 http://orcid.~org/0009 0003-4909-6058 Pittock Hanna 3 Watkins Kate 3 York Brittany~1 and N.~P. S. W.~X. 11, ``Ai-readi: rethinking ai data collection, preparation and sharing in diabetes research and beyond,'' \emph{Nature metabolism}, vol.~6, no.~12, pp. 2210--2212, 2024.

\bibitem{Baxter2024AIREADIRA}
S.~L. Baxter, V.~R. de~Sa, K.~S. Ferryman, P.~Jain, C.~S. Lee, J.~Li-Pook-Than, T.~Y.~A. Liu, J.~P. Owen, B.~Patel, Q.~Yu, L.~M. Zangwill, A.~Bahmani, C.~G. Chute, J.~C. Edberg, S.~Hurst, H.~Ishikawa, A.~Y. Lee, G.~McGwin, S.~K. McWeeney, C.~Nebeker, C.~Owsley, S.~J. Singer, R.~Adib, M.~Adibuzzaman, A.~Alavi, C.~Ashley, A.~Baer, E.~Benton, M.~Blazes, A.~Cohen, B.~A. Cordier, K.~Crist, C.~Cuddy, A.~Gasimova, N.~Gim, S.~S. Hong, T.~Kim, W.-C. Lin, J.~Mitchell, C.~Ngadisastra, V.~Patronilo, J.~Shaffer, S.~Soundarajan, K.~Zhao, C.~Drolet, A.~Lucero, D.~S. Matthies, H.~Pittock, K.~Watkins, B.~York, C.~E. Amankwa, M.~Bangudi, N.~Haboudal, S.~Hallaj, A.~Heinke, L.~Huang, F.~G.~P. Kalaw, A.~Karsolia, H.~Khazaei, M.~Mohammed, K.~U. Simpkins, and X.~Wang, ``Ai-readi: rethinking ai data collection, preparation and sharing in diabetes research and beyond,'' \emph{Nature Metabolism}, vol.~6, pp. 2210 -- 2212, 2024.

\bibitem{GarminStressFAQ}
Garmin, ``What do the stress level numbers mean?'' \url{https://support.garmin.com/en-US/?faq=WT9BmhjacO4ZpxbCc0EKn9}, accessed: 2025-12-02.

\bibitem{Hamman2023RobustCE}
F.~Hamman, E.~Noorani, S.~Mishra, D.~Magazzeni, and S.~Dutta, ``Robust counterfactual explanations for neural networks with probabilistic guarantees,'' in \emph{International Conference on Machine Learning}, 2023.

\bibitem{Guo2021CounterNetET}
H.~Guo, T.~H. Nguyen, and A.~Yadav, ``Counternet: End-to-end training of prediction aware counterfactual explanations,'' \emph{Proceedings of the 29th ACM SIGKDD Conference on Knowledge Discovery and Data Mining}, 2021.

\bibitem{Guidotti2022CounterfactualEA}
R.~Guidotti, ``Counterfactual explanations and how to find them: literature review and benchmarking,'' \emph{Data Mining and Knowledge Discovery}, vol.~38, pp. 2770 -- 2824, 2022.

\end{thebibliography}

\end{document}